\documentclass[sigconf]{acmart} 
\AtBeginDocument{%
  }

\copyrightyear{2023} 
\acmYear{2023} 
\setcopyright{acmlicensed}\acmConference[MM '23]{Proceedings of the 31st ACM International Conference on Multimedia}{October 29-November 3, 2023}{Ottawa, ON, Canada}
\acmBooktitle{Proceedings of the 31st ACM International Conference on Multimedia (MM '23), October 29-November 3, 2023, Ottawa, ON, Canada}

\acmSubmissionID{2425}

\begin{document}

\title{TTPOINT: A Tensorized Point Cloud Network for Lightweight Action Recognition with Event Cameras}


\author{Hongwei Ren}
\authornote{Both authors contributed equally to this research.}
\affiliation{%
  \institution{MICS Trust, \\ The Hong Kong University of Science and Technology (Guangzhou)}
  \city{Nansha,Guangzhou}
  \country{China}}
\email{hren066@connet.hkust-gz.edu.cn}

\author{Yue Zhou}
\authornotemark[1]
\affiliation{%
  \institution{MICS Trust, \\ The Hong Kong University of Science and Technology (Guangzhou)}
  \city{Nansha,Guangzhou}
  \country{China}}
\email{yzhou833@connet.hkust-gz.edu.cn}

\author{Haotian Fu}
\affiliation{%
  \institution{MICS Trust, \\ The Hong Kong University of Science and Technology (Guangzhou)}
  \city{Nansha,Guangzhou}
  \country{China}}
\email{hfu373@connet.hkust-gz.edu.cn}

\author{Yulong Huang}
\affiliation{%
  \institution{MICS Trust, \\ The Hong Kong University of Science and Technology (Guangzhou)}
  \city{Nansha,Guangzhou}
  \country{China}}
\email{yulonghuang@hkust-gz.edu.cn}

\author{Renjing Xu}
\affiliation{%
  \institution{MICS Trust, \\ The Hong Kong University of Science and Technology (Guangzhou)}
  \city{Nansha,Guangzhou}
  \country{China}}
\email{renjingxu@ust.hk}

\author{Bojun Cheng}
\authornote{Corresponding author.}
\affiliation{%
  \institution{MICS Trust, \\ The Hong Kong University of Science and Technology (Guangzhou)}
  \city{Nansha,Guangzhou}
  \country{China}}
\email{bocheng@ust.hk}






\renewcommand{\shortauthors}{Hongwei Ren et al.}

\begin{abstract}
Event cameras have gained popularity in computer vision due to their data sparsity, high dynamic range, and low latency. As a bio-inspired sensor, event cameras generate sparse and asynchronous data, which is inherently incompatible with the traditional frame-based method. Alternatively, the point-based method can avoid additional modality transformation and naturally adapt to the sparsity of events. Still, it typically cannot reach a comparable accuracy as the frame-based method. We propose a lightweight and generalized point cloud network called TTPOINT which achieves competitive results even compared to the state-of-the-art (SOTA) frame-based method in action recognition tasks while only using 1.5\% of the computational resources. The model is adept at abstracting local and global geometry by hierarchy structure. By leveraging tensor-train compressed feature extractors, TTPOINT can be designed with minimal parameters and computational complexity. Additionally, we developed a straightforward downsampling algorithm to maintain the spatio-temporal feature. 
In the experiment, TTPOINT emerged as the SOTA method on three datasets while also attaining SOTA among point cloud methods on all five datasets. Moreover, by using the tensor-train decomposition method, the accuracy of the proposed TTPOINT is almost unaffected while compressing the parameter size by 55\% in all five datasets. 
\end{abstract}




\maketitle

\section{Introduction}
Dynamic Vision Sensors (DVS) offer an attractive option for deploying lightweight networks that can perform real-time action recognition. 
Unlike conventional cameras that sample light intensity at fixed intervals, each DVS pixel independently responds to in situ gradient light intensity changes, signaling the corresponding point's movement in three-dimensional space \cite{lichtsteiner2008128}. This unique capability allows DVS cameras to capture sparse and asynchronous events that respond rapidly to changes in brightness. DVS cameras exhibit remarkable sensitivity to rapid light intensity changes, detecting log intensity in individual pixels and achieving event outputs typically in the order of a million events per second (Meps). In comparison, common cameras can only capture 60 frames per second, highlighting the unparalleled speed of DVS cameras \cite{yao2021temporal}. Moreover, the unique operating principle of DVS cameras grants them superior power efficiency, low latency, less redundant information, and high dynamic range. As such, DVS cameras have found diverse applications in high-speed target tracking, simultaneous localization and mapping (SLAM) \cite{ren2021atfvo}, and industrial automation.

\begin{figure}[h]
\centerline{\includegraphics[width= 8cm]{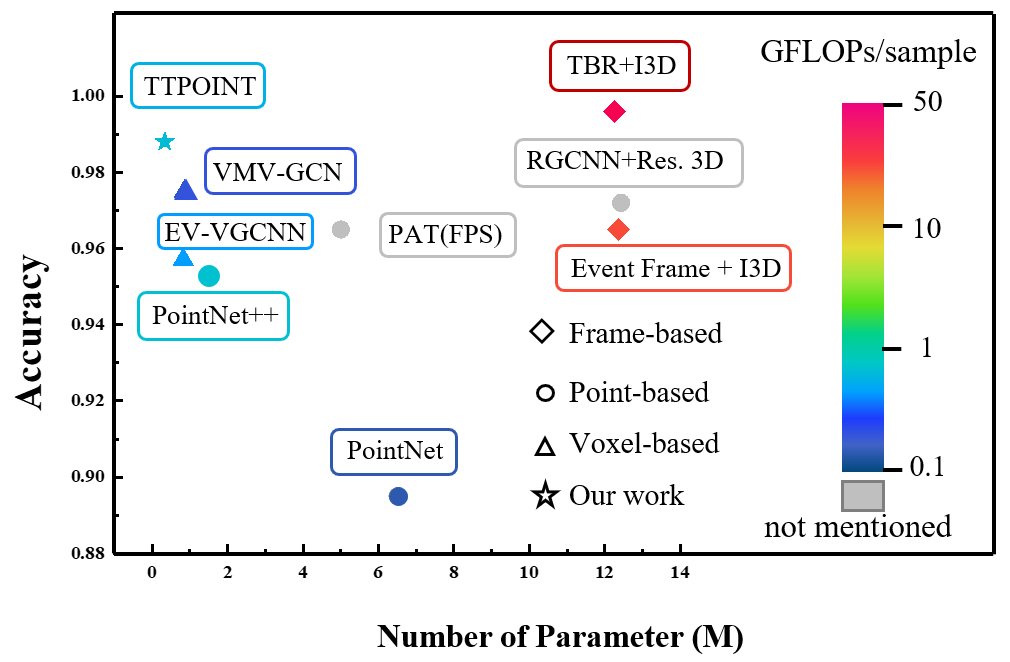}}
\caption{The gain in DVS128 Gesture of our work is visualized. TTPOINT achieves SOTA in the point-based method and maintains very high efficiency. The shapes utilized to illustrate the results correspond to different input data formats, with varying colors denoting FLOPs.
}
\label{fig}
\end{figure}
Event stream contains information on the coordinates, time, and polarity of the event occurrence \cite{posch2014retinomorphic}. The temporal stream can distinguish fast behaviors within several microseconds \cite{gallego2020event}. Several studies have endeavored to extend and improve the problem of processing dynamic motion information from DVS. As shown in Figure 2 (a), the event stream data is segmented into small chunks, and the event for each time period will be mapped and accumulated into the image plane \cite{gehrig2019end}. The grayscale on the plane will be stated as the intensity of the event occurrence at that point. 
However, the framed-based method ignores the fine-grained temporal information within a frame, and the accumulated image gets blurred in fast-moving action scenarios. Moreover, this approach increases the data density and processing volume, diluting the sparse data benefits of an event-based camera. As a result, the frame-based method is unsuitable for scenarios with fast-moving actions and cannot be accomplished with limited computational resources.

In order to overcome the issues discussed above, alternative approaches, such as the point-based method have been proposed. This method is well-suited for analyzing DVS data as each event can be treated as a 3D point with coordinates $(x, y, t)$, which can then be directly fed into the point cloud network, as shown in Figure 2 (b). By doing so, the data conversion process is greatly simplified, and the fine-grained temporal information is preserved \cite{wang2019space}. However, it is worth noting that most existing efforts have been focused on optimizing the network structures, while relatively little attention has been paid to sampling points. As only a small fraction of all points (typically 0.1\% to 1\%) are sampled in the point cloud, it is important to ensure that they are sampled in a careful manner to avoid losing valuable information and to maintain high accuracy in action recognition. Since the point density of DVS events can vary over time, random sampling may not capture enough points in regions with low temporal point density, adversely affecting the capture of complete temporal information of actions.

\begin{figure}[h]
\centerline{\includegraphics[width= 8cm]{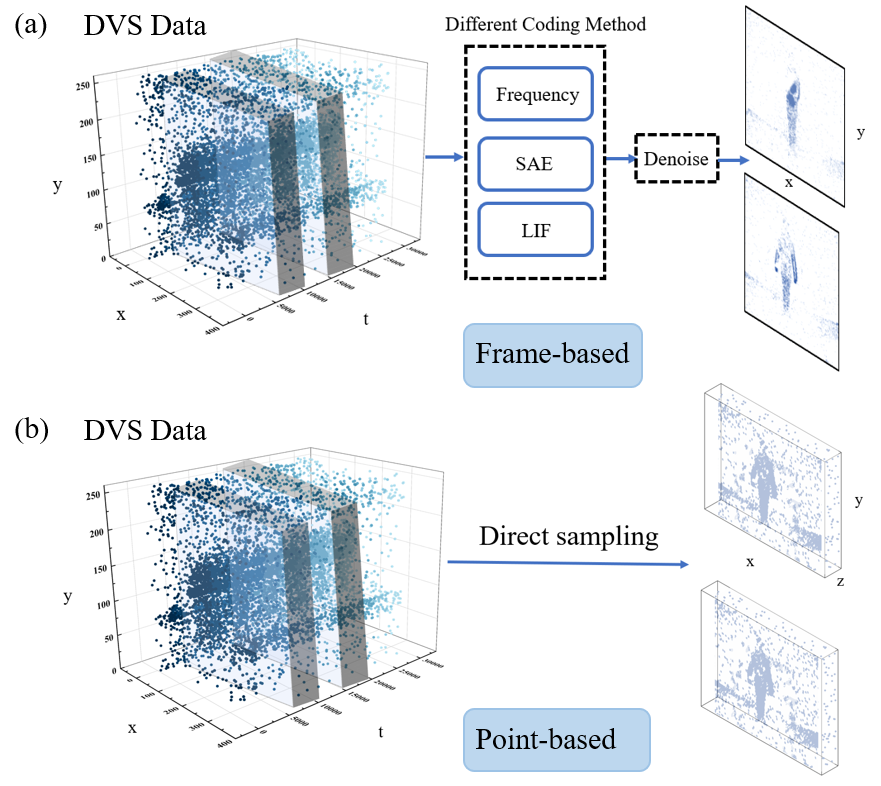}}
\caption{Two different methods of handling event streams, are $(a)$ frame-based methods and $(b)$ point-based methods. Frame-based methods compress a period of events into a grayscale image by SAE, LIF, etc. 
}
\label{fig}
\end{figure}

Our primary goal is to democratize computer vision technology by making it accessible to a wider range of devices and applications. 
It is indeed accurate that conventional action recognition generally displays a negligible processor power consumption. However, with the advent of DVS technology, the power consumption of sensors has been drastically reduced by 90\%-95\%, rendering it much lower than that of conventional ANN processors. Consequently, decreasing the algorithm's power consumption has the potential to significantly reduce overall power consumption and enable the deployment of numerous battery-powered or resource-constrained devices, such as mobile phones, drones, and IoT devices.
To achieve this goal, we introduce a novel tensorized point cloud network, named TTPOINT, for event-based action recognition tasks, as shown in Figure 3. TTPOINT is characterized by its lightweight parameterization and computational demands, making it ideal for resource-constrained devices. 
It features a spatio-temporal prepossessing, a hierarchy structure composed of the tensor-train decomposition module, and a classifier. First, the event stream is divided into many sliding windows of fixed length. For each sliding window, we sample points from $d$ subwindows to ensure the consistency of actions in time. Then, the sampled points are fed into a hierarchy structure to extract point cloud features. The hierarchy structure composes of a tensor-train decomposition module which makes up of residual local extractors and residual global extractors. Finally, the general classifier of Multi-Layer Perceptron (MLP) is used for action recognition. The main contributions of this work are: 
\begin{itemize}
    \item We propose a lightweight point-based network with a generalization ability to handle different-size event-based tasks.
    \item We affiliate the temporal information by subwindow sampling and obtain spatial features by the residual extractor, forming a spatio-temporal network structure.     
    \item We utilize tensor-train decomposition to make the architecture lighter and more efficient with acceptable loss. 
\end{itemize}
\section{Related Work}
\subsection{Frame-based Method}
\begin{figure*}[h]
\centerline{\includegraphics[width= 18 cm]{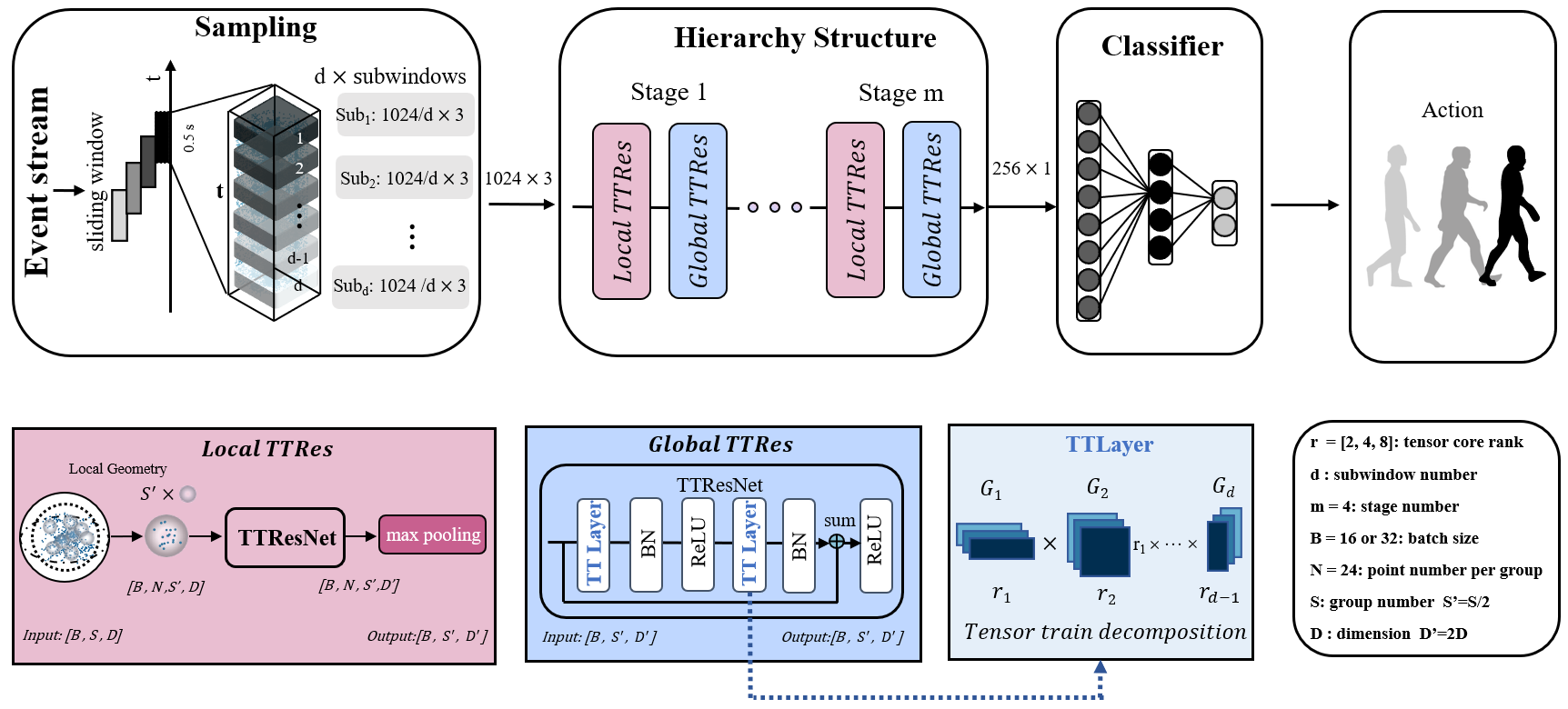}}
\caption{TTPOINT achieves action recognition by processing the event stream through several modules, namely sampling, hierarchy structure, and classifier. Specifically, $LocalTTRes$ is in charge of extracting local geometric features, whereas $GlobalTTRes$ serves to increase the dimensionality of the extracted features and abstract more high-level global features.}
\label{fig}
\end{figure*}
Frame-based methods have been extensively employed for event data processing. These methods convert events into a density map by integrating the number of events within a specific time interval, which can be visualized as a 2D image \cite{innocenti2021temporal}. The resulting image can be directly fed into traditional feature extraction techniques such as Convolutional Neural Networks (CNNs) \cite{amir2017low}, RGCNNs \cite{bi2020graph}, and Asynchronous Sparse Convolutional Networks \cite{messikommer2020event}, all of which have demonstrated high accuracy when using event input, as shown in Figure 1 for benchmarking purposes. However, these methods suffer from several limitations. For instance, the processed frame sizes are typically larger than those of the original event stream data, resulting in high computational costs and large model sizes. Furthermore, the aggression process introduces significant latency \cite{gehrig2019end}, which hinders the use of event cameras in real-time human-machine interaction applications.
\subsection{Point-based Method}
The emergence of PointNet \cite{qi2017pointnet}  allows the point cloud to be processed directly as input. PointNet++ \cite{qi2017pointnet++} adds the Set Abstraction (SA) module to determine the hierarchy architecture of global and local processing of the point cloud. Since the feature extractor of PointNet++ is still using simple MLP, more new methods have been added to improve the quality of point cloud methods. PointConv \cite{wu2019pointconv} uses deep convolution neural networks and deconvolution networks for point cloud processing. Transformer is improved to be more suitable for point cloud tasks and form the architecture of PointTransformer \cite{zhao2021point}. Recently, simple methods also achieve outstanding results. For example, PointMLP \cite{ma2022rethinking} leverages simple ResNet block and affine module and achieves SOTA results in point classification and segmentation tasks.

Applying the point-based method to the event stream requires processing the temporal information without loss of representation with x and y \cite{sekikawa2019eventnet}. ST-EVNet \cite{wang2019space} first achieved this and applied PointNet++ to complete the task of gesture recognition. PAT \cite{yang2019modeling} improved the model and achieved better performance in DVS128 Gesture by combining the self-attention mechanism and Gumbel subset sampling. 
However, these methods rely on random sampling, which fails to capture enough points in low temporal point density regions. Moreover, their models are not optimized for size and computational efficiency and cannot meet the demands of edge devices based on ARM-cortex-M \cite{xie2022vmv}.
Our approach builds upon the strengths of the point cloud, namely its sparse nature and low computational requirements. In contrast to prior methods, we propose a novel sampling algorithm that can capture events occurring at different temporal scales and varying movement speeds. To cater to real-time applications, we optimize the use of lightweight and sparse point-based methods and compress the model while maintaining high performance. Through a comprehensive evaluation of five action recognition datasets, we demonstrate the superiority of our approach over most frame-based methods in terms of accuracy, parameter size, and operational efficiency.

\section{Method}
Our approach builds on the strengths of the point-based method while enhancing the efficiency and effectiveness of the overall algorithm. We achieve this through an action-friendly sampling method. 
To achieve optimal lightweight design, we employ a hierarchical structure for the network, which is then further optimized and compressed to achieve a favorable balance between accuracy and network size. 
\begin{figure*}[!h]
\centerline{\includegraphics[width=18cm]{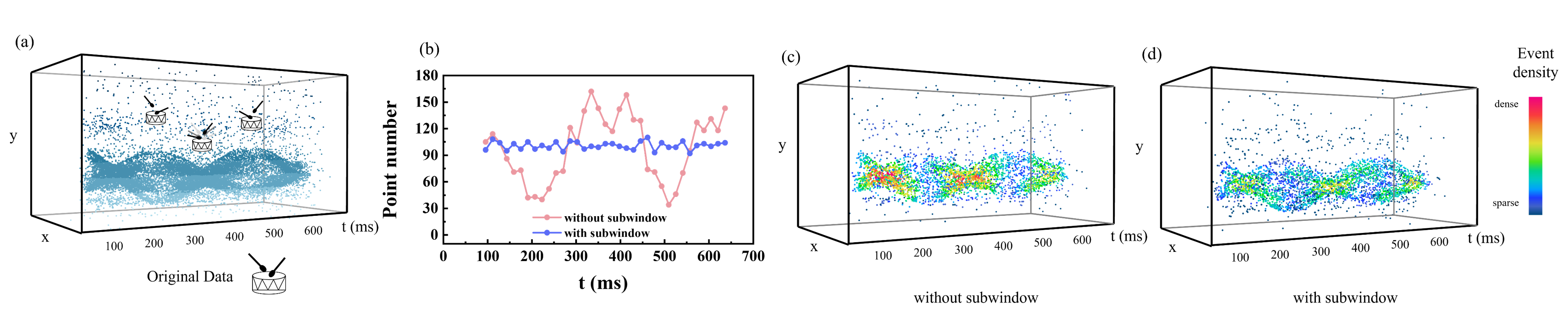}}
\caption{
Visualization results on air drums action with and without subwindow. We visualize the dataset using different colors to denote the density of points, where red represents high density, and blue represents low density. The results are presented in various slices, with (a) showing the original point cloud of the air drums' action. We use statistical analysis to describe the change in point cloud density in the local range, with and without the subwindow, as shown in (b). We also present the point cloud after sampling without the subwindow in (c) and after sampling with the subwindow in (d).
}
\label{fig}
\end{figure*}
\subsection{Event stream}
Event streams are time-series data that record spatial intensity changes of images in chronological order. A simple event can be expressed in $e_m=(x_m,y_m,t_m,p_m)$, where m stands for the index to identify the m-th element in the sequence. And the set of events in a single action recognition $AR$ data can then be expressed as:
\begin{align}
    AR_{raw} = \left\{e_m=(x_m,y_m,t_m,p_m) \mid m=1, \ldots, n\right\}
\end{align}

To apply the point cloud method, four-dimensional $e_m$ has to be simplified into a three-dimensional space-time event. A straight way to do this is to convert $t_m$ into $z_m$:

\begin{align}
    AR_{point}=\left\{e_{m}=\left(x_{m}, y_{m}, z_{m}\right) \mid m=1,2, \ldots, n\right\}
\end{align}%
With $z_m = \frac{t_m - t_1}{t_n-t_1} \mid m\in (1,n)$, $t_1$ and $t_n$ are the beginning and end timestamp of sliding windows, respectively. And $x_m$, $y_m$, $z_m$ are normalized between $[0,1]$.

The sample is divided into sliding windows for two reasons. First, as shown in Table 1, the length of each sample in the action recognition dataset varies. Using the sliding window can normalize the sampling range within each sliding window. Second, the movement is repeated within a sample in most datasets, and the sliding window can isolate a single action from a set of repeated actions within the whole sample.
\begin{align}
    AR_{clip} = clip_i\left \{ e_{k\longrightarrow l}  \right \} \mid i\in (1,n_{win})  \mid  t_l-t_k=L
\end{align}
where $L$ is the length of the sliding window, and $n_{win}$ is the number of the sliding window. $k$ and $l$ are the timestamps of the starting and ending points of the i-th sliding window, respectively.

As previously stated, random sampling within the sliding window is straightforward and is widely used, but it has its limitations. The drawback is illustrated in Figure 4 with a 0.5 s slice window of air drum movement of the IBM DVS128 gesture dataset. As Figure 4 (b) shows, DVS yields a high density of events during rapid movement and a low density of events during slow movement. Therefore, the conventional random sampling method will get points from the moment with high movement speed, and slow movement is underrepresented. This can be seen clearly in 1024 randomly sampled points in Figure 4 (c).

To tackle this, we propose a further division strategy to ensure the capture of action events equally in the time domain. A millisecond-level subwindow is introduced: 
\begin{align}
  clip = \left \{ sub_i\left \{  e_{a\longrightarrow b }\right \}     \mid i\in[0,n_{sub}] \mid t_b -t_a =L_{sub} \right\}   
\end{align}
where $L_{sub}$ is the length of the subwindow, and $n_{sub}$ is the number of the subwindow. So when events are sampled within a 0.5 s slice, the sampling point of each subwindow is the number of points ($N$) divided by $n_{sub}$. As such, the points are more equally sampled over time and guarantee even slow movement can be represented, as illustrated in Figure 4 (b), (d). In the next section, we will list the experimental results and compare the impact of different sampling points on performance.
\subsection{Network}
In this section, we provide a detailed exposition of the network architecture and present a formal mathematical representation of the model.

When we get a series of 3D data $clip_{sample}$ whose dimension is $[1,1024,3]$ from the event stream through the above sampling method, we can input the data into the uncompressed model for point cloud feature extraction. 
We adopt the residual structure in the event processing field, which is proven to achieve excellent results in the point cloud field \cite{ma2022rethinking}. Next, we will introduce the design of the feature extractor for the uncompressed model.
\begin{align}
    \mathcal{P}_i=Max(LocalRes(S_{i-1}))
\end{align}
\begin{align}
    S_i = GlobalRes(\mathcal{P}_i)
\end{align}
\begin{align}
    F = S_m,\quad S_0=clip_{sample}, where\quad i \in (1,m) 
\end{align}
where $F$ is the extracted point cloud feature, which can be sent into the classifier to complete the task of action recognition. The uncompressed model has the ability to extract global and local features and has a hierarchy structure \cite{qi2017pointnet++}. And the dimension of $LocalRes$ input $\in$ is [B, S, D], $m$ is the number of stages for hierarchy structure, $Max$ is the operation of max-pooling, $\mathcal{P}$ is the output of $LocalRes$, and the dimension of $GlobalRes$ input $\in$ [B, S', D'], as illustrated in Figure 3. $B$ is the batch size, $N$ is the point number of a group, $D$ is the dimension of data, and $S$ is the number of groups in overall point cloud data. 
\begin{figure}[h]
\centerline{\includegraphics[width= 8 cm]{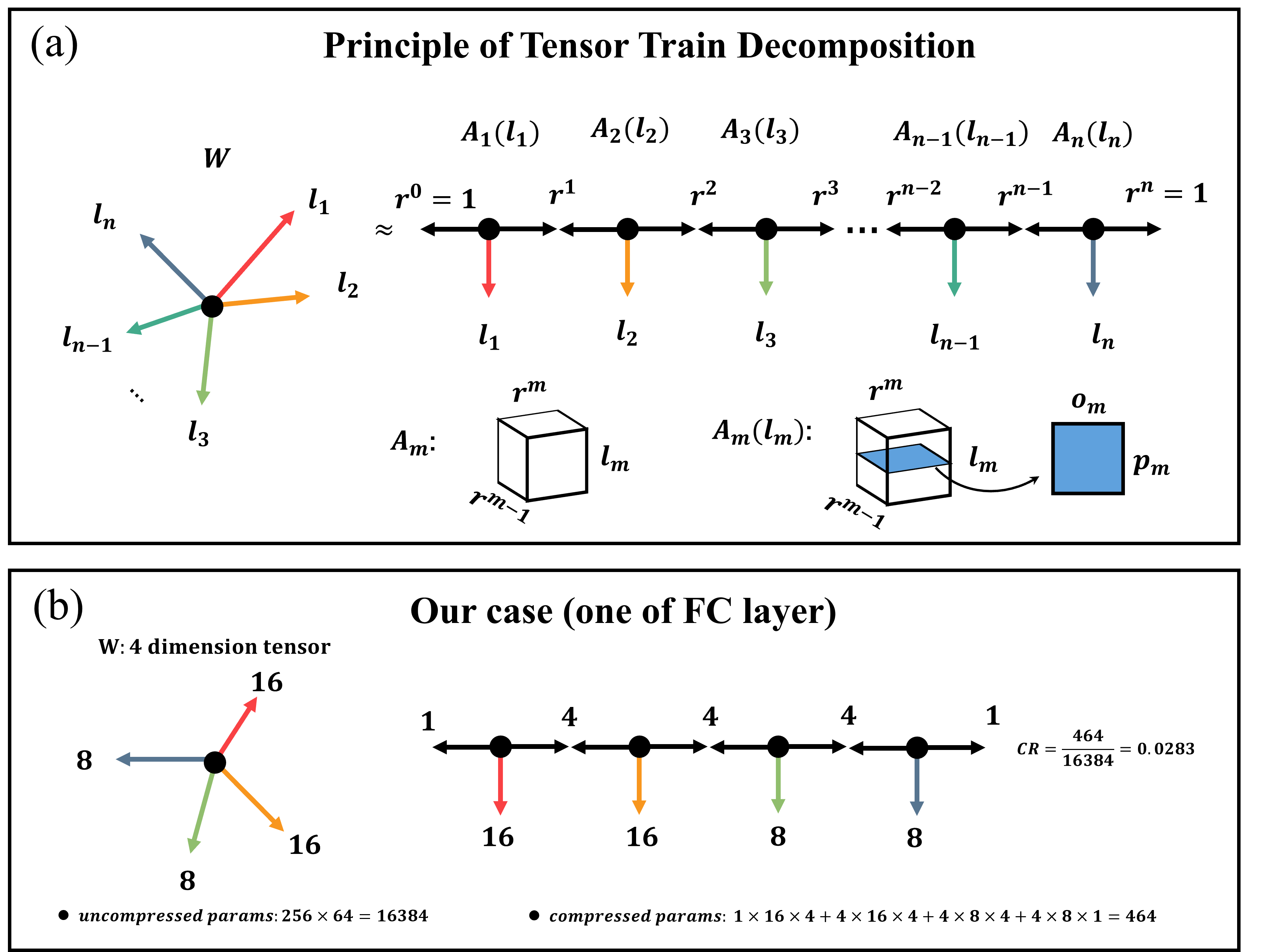}}
\caption{The tensor train decomposition of a $n$-dimensional tensor $\mathcal{W}$ model. Each black dot represents a tensor core, and the lines around the tensor core represent the dimensions.}
\label{fig}
\end{figure}
From a macro perspective, $LocalRes$ is mainly used to extract the local features of the spatial domain and operate the local through the farthest point sampling and grouping method. $GlobalRes$ mainly synthesizes the extracted local features and establishes the global relationship.

Stacking Residual blocks can improve the learning ability of the model, but for sparse event-based data, how to trade off the parameters, computation, and accuracy is a key issue. Compared with the frame-based method, the advantage of point cloud processing is that the amount of data required is small, and the characteristics of events can be utilized. If the point cloud network with similar parameters and computation of frame-based method, it will lose the superiority of using the point cloud to process events. The Residual block can be written as a formula:
\begin{align}
    M(x) = BN(MLP(x))
\end{align}
\begin{align} 
    Res= \mathcal{F}(x+M(\mathcal{F}(M(x)))) 
\end{align}
 where $\mathcal{F}$ is a nonlinear activation function ReLu, $x$ is the input features, $BN$ is batch normalization, and $MLP$ is the muti-layer perception module.

\subsection{Tensor train decomposition}

To achieve the necessary levels of lightweight to surpass even the most advanced voxel-based methods and enable real-time processing on resource-constrained devices, tensor decomposition techniques hold great promise\cite{oseledets2011tensor}. Among these, the tensor-train decomposition (TTD) framework has several notable advantages over more conventional approaches such as Canonical Polyadic (CP) decomposition and Tucker decomposition. 

Firstly, TTD is highly efficient at handling high-dimensional tensors, reducing storage complexity from $O(n^d)$ to $O(d\cdot n \cdot r^2)$ and computational complexity from $O(n^{2d})$ to $O(d\cdot n \cdot r^3)$ (where $n$ is the dimension of the tensor, $d$ is the order of the tensor, and $r$ is the rank size).
Secondly, TTD requires only setting the rank size to balance accuracy and model size, whereas CP decomposition requires specifying the shape of each tensor and introducing additional hyperparameters. Thirdly, the size of each decomposed tensor kernel is uniform in TTD, while Tucker decomposition features a super-large core tensor, posing a challenge for edge computation. Moreover, the feature extractor in our proposed TTPOINT network is implemented using multilayer perceptrons (MLPs), which is a perfect match for the TTD framework. Our experiments confirm the superior performance of TTPOINT leveraging TTD in event-based action recognition tasks.

We use the tensor-train decomposition method to compress the Res module, which will be described in detail below. A $n$-dimensional tensor $\mathcal{W}\in$  $R^{l_{1}\times l_{2} \times ... \times l_{n}}$ can be approximated by a number of tensor cores $\mathcal{A}_{m} \in R^{r_{m-1}\times l_{m}\times r_{m}}$($m \in [1,n]$), where $\mathcal{W}(l_{1}, l_{2},...,l_{n})$ is an element specified by the indices $l_{1}, l_{2},...,l_{n}$, expressed as follows:
\begin{equation}
\mathcal{W}(l_{1}, l_{2},...,l_{n}) = \mathcal{A}_{1}(l_{1})\mathcal{A}_{2}(l_{2})...\mathcal{A}_{}(l_{n})
\end{equation}
$\mathcal{A}_{m}(l_{m}) \in R^{r_{m-1}\times r_{m}}$ is a matrix slice from the 3-dimensional tensor $\mathcal{A}_{m}$ and $r_{m}$ is the rank of $\mathcal{A}_{m}$. For each integer $l_{m}$, it can be further decomposed as $l_{m} = o_{m}\times p_{m}$, where $o$ and $p$ represent the width of the two-layer neural network. Considering the example illustrated in Figure 5 (b) with $l$=16384, the original model has been reshaped into a four-dimensional tensor with the weight, denoted as $l_m$=[16,16,8,8], meaning 16$\times$16$\times$8$\times$8=16384. Likewise, the model with  $o$=256 and $p$=64 also been reshaped in $o_m$=[4,4,4,4], and $p_m$=[4,4,2,2]. As a result, each tensor core $\mathcal{A}_{m}$ can be transformed and reformed into $\mathcal{A}_{m}^{*}\in R^{r_{m-1}\times r_{m}\times o_{m}\times p_{m}}$. The double-index trick, as such is the core function to decompose the fully-connected computation in MLP cells.
\begin{equation}
\begin{split}
& \mathcal{W}((o_{1},p_{1}), (o_{2},p_{2}),...,(o_{n},p_{n})) \\
&=\mathcal{A}_{1}^{*}(o_{1},p_{1})\mathcal{A}_{2}^{*}(o_{2},p_{2})...\mathcal{A}_{n}^{*}(o_{n},p_{n})
\end{split}
\end{equation}
The whole compression process of $\mathcal{W}$ is illustrated in Figure 5 (a), and (b) represents a specific scenario with unique data to facilitate comprehension. If they are in CNN, they represent the height and width of the image, and if they are in the point cloud processing, they represent the input and output dimension values. In general, the large-scale matrix-vector multiplication is the most expensive computation in the MLP cell, generically denoted as:
\begin{equation}
\mathcal{Y} = \mathcal{W} \mathcal{X} + \mathcal{B}
\end{equation}
 To approximates $\mathcal{W} \mathcal{X}$ with much fewer parameters, the weight matrix $\mathcal{W}$ is reshaped into a tensor $\mathcal{W}\in$ $R^{(o_{1}\times o_{2}\times ...\times o_{n})\times (p_{1}\times p_{2} \times ...\times p_{n})}$. Similarly, $\mathcal{X}$ and $\mathcal{B}$ can be reshaped into $n$-dimentional tensors $\mathcal{X}\in R^{l_{1}\times l_{2} \times ...\times l_{n}}$ and $\mathcal{B}\in R^{l_{1}\times l_{2} \times ...\times l_{n}}$. As a result, the output $y$ can also become a $n$-dimentional tensor $\mathcal{Y}\in R^{l_{1}\times l_{2}\times ...\times l_{n}}$. Therefore, the matrix-vector multiplication in MLP cells can be reformulated as follows: 
 \begin{equation}
\begin{split}
&\mathcal{Y} (l_{1}, l_{2},...,l_{n}) = \sum \nolimits...\sum \nolimits [\mathcal{A}_{1}^{*}(o_{1},p_{1}) \\ & \mathcal{A}_{2}^{*}(o_{2},p_{2})
...\mathcal{A}_{n}^{*}(o_{n},p_{n})\mathcal{X}] + \mathcal{B}(l_{1}, l_{2},...,l_{n})
\end{split}
\end{equation}
Due to the tensor decomposition strategy, the computational complexity in the tensor-compressed MLP decreases a lot. For the basic uncompressed model mentioned above, after compression, the model size is compressed by 55\% to merely 0.334 M, and the number of floating-point operations is reduced to 0.587 GFLOPs by 46\%. In TTPOINT, we replace the MLP in Res block with TTlayer which is compressed MLP by the formula: 
\begin{equation}
    \mathcal{T}(x) = BN(TTlayer(x))
\end{equation}
\begin{equation}
TTRes= \mathcal{F} (x+\mathcal{T}(\mathcal{F} (\mathcal{T}(x))))
\end{equation}
\begin{equation}
    \mathcal{P}_i =Max(LocalTTRes(ST_{i-1}))
\end{equation}
\begin{equation}
    ST_i = GlobalTTRes(\mathcal{P}_i)
\end{equation}
\begin{equation}
    TTPOINT = ST_m, \quad ST_0 = clip_{sample}, \quad where \quad i \in (1,m)
\end{equation}
where the rank of TTlayer will provide in the next section. A lite point cloud network with tensor train decomposition is set up.

\begin{table*}
\centering
\renewcommand\arraystretch{1}
\scalebox{0.9}{
\begin{tabular}{ccccccccc}
\hline
Dataset            & Classes       & Sensor        & Resolution         & Avg.length(s) & Train Samples & Test Samples &Sliding window  &Overlap\\ \hline
Daily DVS \cite{liu2021event}          &12            &  DAVIS128              &  128x128              &6   &2924 &731 &1.5s &0.5s\\
DVS128 Gesture \cite{amir2017low}    & 10/11         & DAVIS128      & 128x128                 & 6.52          & 26796         & 6959 &0.5s &0.25s\\
DVS Action \cite{miao2019neuromorphic} & 10            & DAVIS346      & 346x260                & 5             & 912          & 116 &0.5s &0.25s\\
HMDB51-DVS \cite{bi2020graph}        & 51            & DAVIS240      & 320x240                 & 8             & 24463         & 6116  &0.5s &0.5s\\ 
UCF101-DVS  \cite{bi2020graph}        &101            &  DAVIS240             & 320x240             & 6.6  &119214 &29803 &0.5s &0.5s\\\hline
\end{tabular}
}
\caption{Information of different action recognition datasets with event camera.}
\end{table*}

\section{Experiment}
In this section, we train and test the performance of the proposed model on the server platform. The experiment includes the accuracy performance of the model on the datasets, the size of model parameters, and the number of floating-point operations (FLOPS). The specific parameters of our platform are CPU: AMD 7950x, GPU: RTX 4090, and Memory: 32GB.

\subsection{Datasets}
We evaluate TTPOINT on five popular action recognition datasets. The resolution, total classes, and other information of each dataset are shown in Table 1.

\textbf{Daily DVS} dataset \cite{liu2021event} contains 1440 recordings of 15 subjects performing 12 different common daily actions, such as standing up, walking, and carrying boxes. 
Each subject performed each action under the same conditions, and the recordings are all within 6 seconds. 

\textbf{DVS128 Gesture} dataset \cite{amir2017low} is 11 classes of gestures collected by IBM in the actual environment with the iniLabs DAVIS128 camera. The sample resolution of this dataset is 128$\times$128. That is, the coordinate value range of x and y is [1, 128]. 
The dataset includes 1342 examples, divided into 122 experiments that collected 29 subjects under different lighting conditions.

\textbf{DVS Action} dataset \cite{miao2019neuromorphic} is 10 classes of action collected with DAVIS346 camera. The sample resolution of this dataset is 346$\times$260. 
The dataset was recorded in an empty office, and 15 subjects performed 10 different actions. 
The record file is named after each action, such as crossing arms, getting up, kicking, picking up, jumping, sitting down, throwing, turning, walking, and waving.

\textbf{HMDB51-DVS and UCF101-DVS} datasets \cite{bi2020graph} are converted from the dataset collected by traditional cameras, and the conversion mode uses DAVIS240. UCF101 contains a collection of 13,320 videos demonstrating 101 unique human actions, whereas HMDB51 consists of 6,766 videos depicting 51 categories of human actions. The sample resolution of these datasets is 320$\times$240. 

\subsection{Implement Details}
\subsubsection{Pre-processing}
The time set for sliding windows is 0.5 s, 1 s, or 1.5 s. In the DVS128 Gesture dataset and DVS Action dataset, the coincidence area of adjacent windows is set as 0.25 s. In other datasets, this is set to 0.5 s. Since there are many noises in the DVS Action dataset during the actual recording, we use the denoising method to reduce the probability of sampling noise points, and this dataset has no labels. After we are statist and visualize the files, we find that almost all actions occur after half of the total dataset, so when we build the dataset, the sampling start point starts from half of the total timestamps. Compared with the manual annotation method \cite{wang2019space}, our method is slightly rough but still achieves better results on the DVS Action dataset.

\subsubsection{Network Structure}
The rank of tensor train decomposition affects the size of the compressed model and its performance on the dataset, so it is important to select an appropriate rank \cite{man2023ranksearch}. The rank of TTPOINT is 8, and the maximum compression ratio can reach 14. Additionally, we set this value to 4 in the ablation study for intensive compression, and the maximum compression ratio can reach 53, and the parameter quantity reduces by 0.03 M. We made statistics on the width of MLP in the model. Following the change of network parameters, the width is distributed in the set [1024, 512, 256, 128, 64, 32, 16]. So we set to rank as follows according to experience, 1024: [8, 8, 4, 4], 512: [8, 4, 4, 4], 256: [4, 4, 4], 128: [4, 4, 4, 2], 64: [4, 4, 2, 2], 32: [4, 2, 2], 16: [2, 2, 2, 2].  TTPOINT is divided into three modules: preprocess module, feature extractor, and classifier. Consult \cite{ma2022rethinking}, we used four stages of the TTResidual block. The number of k-nearest neighbors for local feature extraction is 24.

\subsubsection{Training Hyperparameter}
Our training model uses the following hyperparameters, Batch Size: 16 or 32, Number of Points: 1024, Optimizer: momentum, Initial Learning Rate: 0.1, Scheduler: cosine, Max Epochs: 350.

\subsection{Results of Action Recognition}
\subsubsection{Daily DVS}
TTPOINT outperformed all methods and achieved an accuracy of 99.1\% on this dataset, as shown in Table 2. However, despite using nearly 80 times more parameters, the frame-based method still yielded poor accuracy. We visualize the event stream in this dataset, which featured high recording quality, minimal noise, and clear actions. By using the point cloud sampling method, we could accurately capture the motion characteristics of actions. Initially, we employed a sliding window of 0.5 s, but this only approximately resulted in an accuracy of 85\%. After adjusting it to 1.5 s, TTPOINT achieved remarkable accuracy of 99.1\%, clearly demonstrating its superiority over other methods.

\begin{table}[]
\centering
\renewcommand\arraystretch{1}
\scalebox{0.85}{
\begin{tabular}{cccc}
\hline
Name              & Param.(x$10^6$) & GFLOPs & Acc    \\ \hline
I3D\cite{carreira2017quo}     & 49.19 &59.28    & 0.962 \\
TANet\cite{liu2021tam}                & 24.8  &65.94    & 0.965 \\
VMV-GCN \cite{xie2022vmv}            & 0.84   &0.33  & 0.941 \\
TimeSformer \cite{bertasius2021space}        & 121.27  &379.7   & 0.906 \\
Motion-based SNN \cite{liu2021event}    & -    & - & 0.903 \\ \hline
\textbf{TTPOINT }        & \textbf{0.335} &\textbf{0.587} & \textbf{0.991} \\ \hline
\end{tabular}
}
\caption{Model's accuracy and complexity on Daily DVS.}
\end{table}

\begin{table}[]
\centering
\renewcommand\arraystretch{1}
\scalebox{0.85}{
\begin{tabular}{cccc}
\hline
Method     & Param.(x$10^6$) & GFLOPs & Acc   \\ \hline
TBR+I3D \cite{innocenti2021temporal}    & 12.25  & 38.82  & 0.996 \\
Event Frames + I3D \cite{bi2020graph} &12.37 &30.11 & 0.965 \\
EV-VGCNN \cite{deng2021ev}   & 0.82   & 0.46 & 0.957 \\
RG-CNN \cite{miao2019neuromorphic}     & 19.46  & 0.79      & 0.961 \\
PointNet++ \cite{wang2019space} & 1.48   & 0.872  & 0.953 \\ 
VMV-GCN \cite{xie2022vmv} & 0.86   & 0.33 & 0.975 \\ \hline
TTPOINT   & \textbf{0.334}  & \textbf{0.587}  & \textbf{0.988}
\\ \hline
\end{tabular}
}
\caption{Model's accuracy and complexity on DVS128 Gesture.}
\end{table}

\subsubsection{DVS128 Gesture}
Table 3 shows the performance and network size of DVS 128 Gesture. The highest accuracy achieved is 99.6\% \cite{innocenti2021temporal}. However, this approach is frame-based and requires a large amount of memory for storage and calculation, which is about 50 times larger than our model. With model compression, TTPOINT has a parameter size of only 0.334 M and a total number of floating-point operations of 0.587 GFLOPs, while still achieving an accuracy of 98.8\% on the DVS128 Gesture dataset. The parameter quantity of our method is greatly reduced, but the FLOPs are not reduced as much, possibly due to more operations on the original basis in the tensor-train method. It's also worth noting that TTPOINT achieved an accuracy of 97.8\% on the sliding window testset. By utilizing a voting method on individual event streams, the classification result with the highest number of votes shows remarkable accuracy across 240 test sequences.

\begin{table}[]
\centering
\renewcommand\arraystretch{1}
\scalebox{0.85}{
\begin{tabular}{ccccc}
\hline
Method     & Param.(x$10^6$) &GFLOPs      & Num. & Acc   \\ \hline
Deep SNN \cite{gu2019stca}   & -   & -   & -    & 0.712 \\
Motion-based SNN \cite{liu2021event} & -   & -  & -    & 0.781 \\
PointNet  \cite{qi2017pointnet} &3.46 &  0.450   & 512  & 0.751 \\
ST-EVNet  \cite{wang2020st}   &1.6 & -   & 1024 & 0.887 \\ \hline
TTPOINT      & \textbf{0.334}  & \textbf{0.587}  & \textbf{1024} & \textbf{0.927} \\ \hline
\end{tabular}
}
\caption{Model's accuracy and complexity on DVS Action.}
\end{table}

\subsubsection{DVS Action}
Table 4 illustrates the performance of the TTPOINT for DVS Action. TTPOINT achieves an accuracy rate of 92.7\% on the DVS Action dataset. However, it can be seen from the number of training and test samples of the dataset that DVS Action is smaller than others. The knowledge learned from the large model may be overfitted, and the representation ability of the compressed model is weakened to a certain extent so that it can perform better on this small dataset. We use the same model architecture on the five datasets and do not design targets for each dataset. This means that we may get better results if we adjust the architecture for different datasets.

\begin{table}[]
\centering
\renewcommand\arraystretch{1}
\scalebox{0.85}{
\begin{tabular}{cccc}
\hline
Method     & Param.(x$10^6$) & GFLOPs & Acc   \\ \hline
C3D \cite{tran2015learning}        & 78.41  & 39.69  & 0.417 \\
I3D \cite{carreira2017quo}      & 12.37  & 30.11  & 0.466 \\
ResNet-34 \cite{he2016deep}  & 63.70  & 11.64  & 0.438 \\
ResNext-50 \cite{hara2018can} & 26.05  & 6.46   & 0.394 \\
ST-ResNet-18 \cite{samadzadeh2020convolutional} & -  & -   & 0.215 \\
RG-CNN \cite{bi2020graph}     & 3.86   & 12.39  & 0.515 \\ \hline
TTPOINT   & \textbf{0.345}  & \textbf{0.587}  & \textbf{0.569} \\ \hline
\end{tabular}
}
\caption{Model's accuracy and complexity  on HMDB51-DVS.}
\end{table}

\subsubsection{HMDB51-DVS}
The first three datasets are a few categories of action recognition datasets, so the HMDB51-DVS dataset is used to verify whether the proposed method can be generalized to large datasets. Compared with the DVS128 Gesture dataset, we found that other methods, which are frame-based or voxel-based must increase the parameters several times or even dozens of times in order to learn the knowledge of this dataset. We can discover from Table 5 that the TTPOINT with the same structure can achieve perfect results. It also can be seen from Table 5 that the number of parameters is not consistently compared with the DVS128 Gesture dataset’s model only because the final classifier’s output has changed from 10 to 51.  Moreover, the TTPOINT can be expanded to accommodate such complicated tasks, and we project a further improvement of the accuracy by increasing the TTPOINT’s size.
\begin{table}[]
\centering
\renewcommand\arraystretch{1}
\scalebox{0.85}{
\begin{tabular}{cccc}
\hline
Name          & Param.(x$10^6$) & GFLOPs & Acc   \\ \hline
C3D \cite{tran2015learning}          & 78.41                         & 39.69  & 0.472 \\
ResNext-50.3D \cite{hara2018can} & 26.05                          & 6.46   & 0.602 \\
I3D \cite{carreira2017quo}          & 12.37                         & 30.11  & 0.635 \\
RG-CNN \cite{bi2020graph}       & 6.95                          & 12.46  & 0.632 \\
ECSNet  \cite{chen2022ecsnet}      & -                             & 12.24  & 0.702 \\ 
Event-LSTM \cite{annamalai2022event} &21.4  &- &0.776 \\ \hline
\textbf{TTPOINT }      & \textbf{0.357}                         & \textbf{0.587}  & \textbf{0.725} \\ \hline
\end{tabular}
}
\caption{Model's accuracy and complexity on UCF101-DVS.}
\end{table}

\subsubsection{UCF101-DVS}
To further validate the generalization and effectiveness of TTPOINT, we conducted tests on a larger dataset with 101 classification results. The performance of TTPONT is shown in Table 6, not surpassing Event-LSTM as the second-best model. After being divided into sliding windows, the trainset of UCF101-DVS reached as high as 1192014, which is a huge challenge for the model. We still haven't changed the model except for modifying the output of the classifier to 101 classes. TTPOINT achieved an accuracy of 72.5\%, and we envision increasing model complexity, which may further improve the accuracy.

In summary, we used the same TTPOINT to conduct a comprehensive evaluation of small, medium and large datasets. We found that TTPOINT is a model with the ability to generalize and achieve very good results on data of different sizes without the need for a targeted design of the network structure. 

\subsection{Ablation Study}
\subsubsection{The time interval of subwindow}
To evaluate the new pre-processing method, we apply four different pre-processing parameters and train the TTPOINT with both the DVS128 Gesture dataset and the DVS Action dataset. The results are summarized in Table 7. We uniformly sample subwindows on the millisecond timesteps axis based on the sliding windows to ensure that the points and actions sampled on the time axis are strongly correlated. The length of the subwindows is set to four different values. We train TTPOINT with four different pre-processing data from the DVS128 Gesture dataset and DVS Action, and one does not use the subwindows method and another three uses. After repeated experiments, the best performance of TTPOINT on the two datasets is obtained. It finds that the accuracy of the model using the subwindow method was nearly 0.8\% higher with 0.978 and 0.970 in DVS128 Gesture, and the accuracy of the model using the subwindow method was near 5.2\% higher with 0.927 and 0.875 in DVS Action.
\begin{table}[]
\centering
\renewcommand\arraystretch{1}
\scalebox{0.9}{
\begin{tabular}{ccccc}
\hline
Subwinow(x$10^3 \mu s$)   &No subwindow &250  & 125    &62.5        \\ \hline
DVS128 Gesture \cite{amir2017low} &0.970 &0.973  & 0.978   &0.975     \\
Improve(\%)   &- &0.3 & \textbf{0.8}    &0.5   \\
DVS Action \cite{miao2019neuromorphic}   &0.875 &0.906 & 0.927 &0.911        \\
Improve(\%)    &- &3.1& \textbf{5.2}    &3.6 \\ \hline
\end{tabular}
}
\caption{ The improvement of accuracy on different lengths of subwindow which is $6.25\times10^4$, $1.25\times10^5$, $2.5\times10^5$ and $5\times10^5$ timestamps.}
\end{table}
In addition, we find that the accuracy of the model decreases as the number of subwindows increases from 125 to 62.5. We infer this phenomenon occurs because too many subwindow sampling distracts the model from the point-dense region, resulting in a dilution of important information. So in different scenarios, the optimized subwindow number may be different which is up to the speeds of the movement.

\subsubsection{The effect with compression}
\begin{table}[]
\centering
\renewcommand\arraystretch{1}
\scalebox{0.8}{
\begin{tabular}{cccc}
\hline
Dataset    & Uncompressed Model & TTPOINT        & TTPOINT\_tiny \\ \hline
Daliy DVS \cite{liu2021event}     & 0.990    & \textbf{0.991} & 0.988        \\
DVS Action \cite{miao2019neuromorphic}    & 0.911    & \textbf{0.927} & 0.885         \\
DVS128 Gesture \cite{amir2017low}& 0.975    & \textbf{0.978} & 0.970         \\ 
HMDB51-DVS \cite{bi2020graph}    & 0.590    & \textbf{0.569} & 0.545         \\ \hline
\end{tabular}
}
\caption{ The effect of performance with tensor-train decomposition. When working with datasets that are small to medium-sized, compression may result in positive outcomes. However, when dealing with larger datasets, compression may be less beneficial and could even lead to negative effects.}
\end{table}

We conducted a comparison between the accuracy of uncompressed models, TTPOINT, and TTPOINT\_tiny, which is mentioned in 4.2.2 with the rank of 4, across three datasets of different sizes, as shown in Table 8. Surprisingly, we found that the accuracy of the model improved on the DVS ACTION, Daily DVS, and DVS128 Gesture datasets, but there were more severe drop points in the HMDB51-DVS dataset. We must emphasize that the models are identical, except for the classifier, across the five different-sized datasets. We suspect that for small datasets, overfitting was prevalent, but the use of tensor train decomposition effectively reduced this phenomenon, as shown by the rising trend in the testset results. 
\begin{figure}[h]
\centerline{\includegraphics[width= 8 cm]{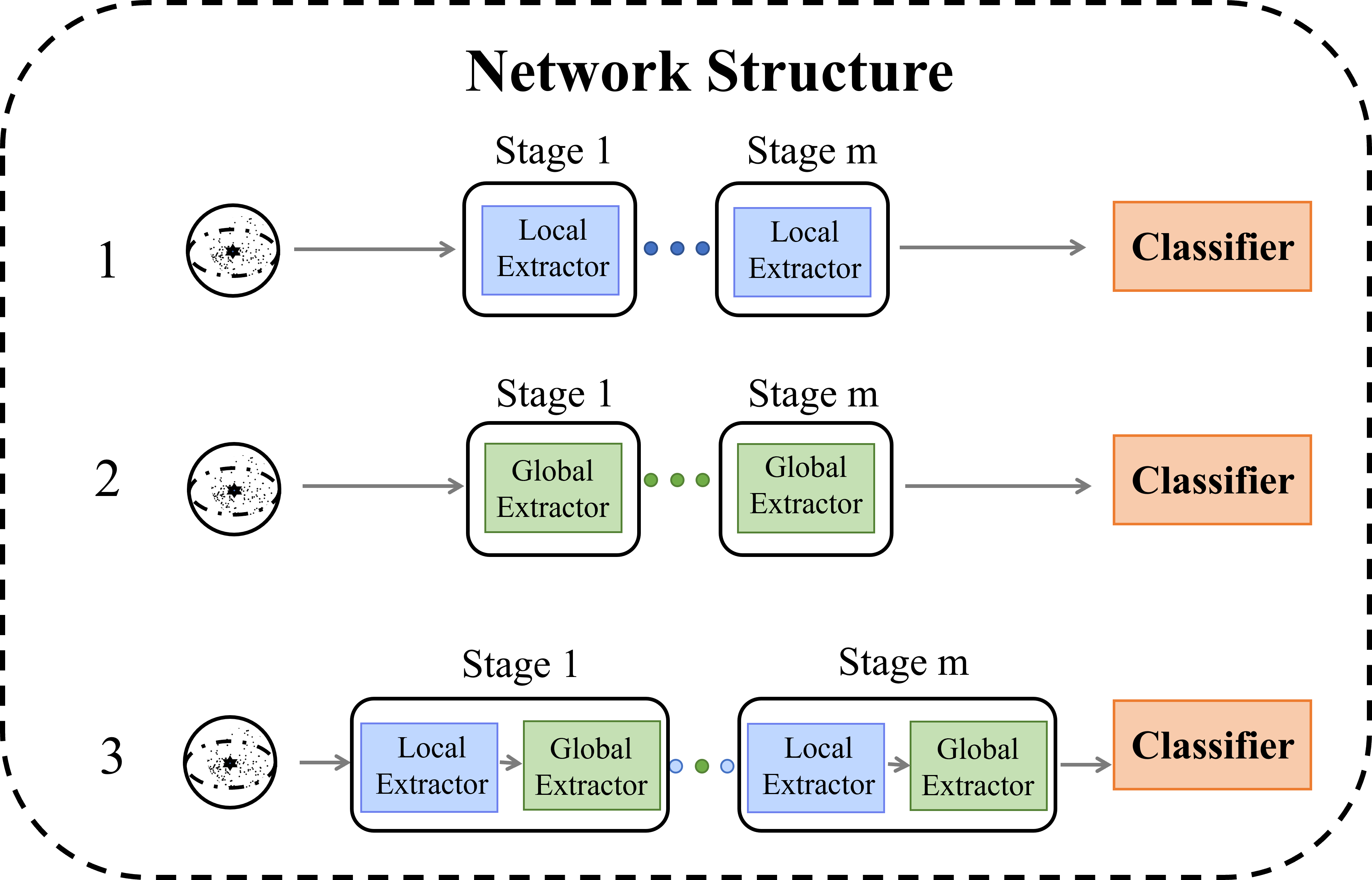}}
\caption{The ablation study on the local and global extractor. Method 1: only with local extractor block; Method 2: only with global extractor block; Method 3: apply both the local global extractor }
\label{fig}
\end{figure}
Conversely, for the HMDB51-DVS dataset, which is significantly larger than the others, we observed underfitting and a substantial loss in accuracy due to excessive compression. We proposed two potential solutions: increasing the rank size of tensor train decomposition or modifying the structure to make it more complex. 
Furthermore, it is important to note that overly aggressive compression can result in suboptimal model performance. This was demonstrated by TTPOINT\_tiny, which showed a decrease in accuracy due to excessive compression. Ultimately, the level of compression that should be applied depends on the desired balance between model accuracy and parameter quantity in practical applications.

\subsubsection{The local and global extractor}
To investigate the contribution of local and global feature extractors in the entire model, we conducted three comparative experiments on DVS128 Gesture as shown in Figure 6. The first experiment group, TTPOINT, lacked global feature extractors, the second group lacked local feature extractors, and the third group had both. Their respective accuracy on the sliding window testset was 96.19\%, 90.2\%, and 97.78\%. These experimental results indicate that the local feature extractor is a crucial component of the model's performance, while the global feature extractor contributes to further improving the model's accuracy. TTPOINT's overall architecture without the local feature extractor is similar to PointNet, which results in poor performance due to the loss of local geometry information in the point cloud. Compared to other point-based methods, TTPOINT has a hierarchical structure similar to PointNet++'s Set Abstraction(SA). However, our model stands out thanks to its lightweight and efficient extractor design, offering not only ultra-high accuracy but also being sufficiently lightweight.

\section{Conclusion}
We propose a lightweight point cloud network named TTPOINT specifically for event camera action recognition, evaluate it on five mainstream corresponding datasets, and obtain excellent results. It performs well in accuracy and is suitable for low-latency applications with limited computational power.

In our future work, we will devote ourselves to the architecture development of the combination of point cloud processing network and spiking neural network (SNN) and can design a neuromorphic computing algorithm that is hardware and software friendly.


\bibliographystyle{ACM-Reference-Format}
\balance
\bibliography{sample-sigconf}

\end{document}